\documentclass[letterpaper,11pt]{easychair}

\usepackage[utf8]{inputenc}
\usepackage{fancyvrb}
\usepackage{color}
\usepackage{amsmath,amsfonts,amsthm,amssymb}
\usepackage{mathrsfs} 
\usepackage{graphicx}
\usepackage{multirow}
\usepackage{fancyhdr}
\usepackage{extarrows}
\usepackage{framed}
\usepackage{tikz}
\usepackage{float}
\usepackage{subfigure}
\usepackage{algorithm}
\usepackage[noend]{algpseudocode}
\usetikzlibrary{arrows}
\usepackage{tabularx}
\usepackage{hyperref}
\usepackage{enumerate}
\usepackage{comment}
\usepackage{ragged2e}
\hypersetup{colorlinks}

\title{Deep Learning Methods for Signature Verification}

\author{
Zihan Zeng\inst{1}
\and
    Jing Tian\inst{2}
}

\institute{
  Beijing Normal University, Zhuhai, 
   Guangdong, China\\
  \email{chinazzh2012@gmail.com}
\and
   Dalian University of Technology,
   Tianjin, China\\
   \email{tj11164@mail.dlut.edu.cn}\\
 }

\titlerunning{Deep Learning Methods for Signature Verification}
\begin{document}

\maketitle

\begin{abstract}
Signature is widely used in human daily lives, and serves as a supplementary characteristic for verifying human identity. However, there is rare work of verifying signature. In this paper, we propose a few deep learning architectures to tackle this task, ranging from CNN, RNN to CNN-RNN compact model. We also improve Path Signature Features by encoding temporal information in order to enlarge the discrepancy between genuine and forgery signatures. Our numerical experiments demonstrate the effectiveness of our constructed models and features representations.

\end{abstract}
\setcounter{tocdepth}{2}
{\small
\tableofcontents}

\section{Introduction}
\label{sect:introduction}

Nowadays, signature is playing an important role in human daily lives, and serves as a supplementary characteristic to facilitate identity verification in many scenarios, e.g., credit card fraud and criminal detections~\cite{Signature2019}. On the contrast, signature forgery also develops rapidly, and deteriorates people's assets severely.   In some regions, signature forgery has even become some illegal industry for profit. On Ebay, there exists services to generate fake signatures~\cite{Forgery2019}. Some people use fake signatures to defraud others' assets, and bilk properties of large organizations such as insurance companies and banks. In the United States, since the real estate bubble burst in 2006, the forged signatures on the documents have caused a \$25 billion settlement between lenders, governments, and most states until 2012~\cite{Brady2012}. And in 2019, the British National Crime Agency and the British Financial Conduct Authority were asked to investigate allegations that bank employees forged signatures on court documents to recover debts and recover homes, which is a scandal~\cite{Vicky2019}.

To avoid abusing forgery signatures, building an effective signature verification method to distinguish genuine signatures from forgery ones is important . The challenges of achieving successful signature verification can be largely evolved into two aspects: (i) intrapersonal variance is high; (ii) interpersonal variance is low. The former one refers to the fact that a single person may sign very differently under different conditions.  The later one means that if the forger has observed the genuine signature, he/she may relatively easily mimic one out with similar geometric properties. The above challenges aggravate the success of accurate verification. To resolve these challenges, state-of-the-art signature verification technique makes use of discrete wavelet transform (DWT) to increase the discrepancy between genuine and forgery signatures, followed by a simple feedforward neural network (consisting of stacking fully connected layers) to do a binary classification~\cite{fahmy2010online}. But the feature extracted by DWT is computationally intensive and not easily implemented properly, which blocks it to abroad usage. 

On the other hand, deep learning has demonstrated remarkable success in many scientific applications~\cite{deng2014deep}, ranging from computer vision to handwritten areas~\cite{ribeiro2011deep}. Particularly, in handwritten areas, based on varied decorated feature extractions, such as Bezier curves and Path Signature Features~\cite{chen2017compact}, Convolutional Neural Network (CNN) and Recurrent Neural Network (RNN) have shown superiority than traditional pattern recognition methods. Therefore, it is natural to ask if utilizing modern deep learning architectures can build a successful signature verification system. To answer this question, in this paper, we propose varying ways of building  signature verification systems used modern deep learning techniques.  The following bullet points summarize our key contributions
\begin{itemize}
	\item We build and investigate varying modern deep learning methods to achieve effective signature verification, ranging from CNN, RNN to CNN-RNN compact models. Our constructed models are easy to implement and deploy comparing with state-of-the-art DWT approach. And we numerically demonstrate the effectiveness of our proposed methods for signature verification. 
	
	\item We improve the Path Signature Feature by encoding temporal information into it to amplify the discrepancy in the temporal aspect to resolve the issue of high interpersonal variance of signatures. 
	
\end{itemize}

The reminder of this paper is organized as follows. At first, in Section~\ref{sect:feature_extraction}, we present how feature extraction works. Then we move on to describe the deep learning model architecture candidates in Section~\ref{sect:nn_methods}. Next, in Section~\ref{sec.numericexp}, we numerically test and compare the performance of our built models, and finally conclude this paper in Section~\ref{sec.conclusion}.




\section{Feature Extraction}
\label{sect:feature_extraction}

\begin{figure}[h]
	\begin{centering}
		\includegraphics[width=1\textwidth]{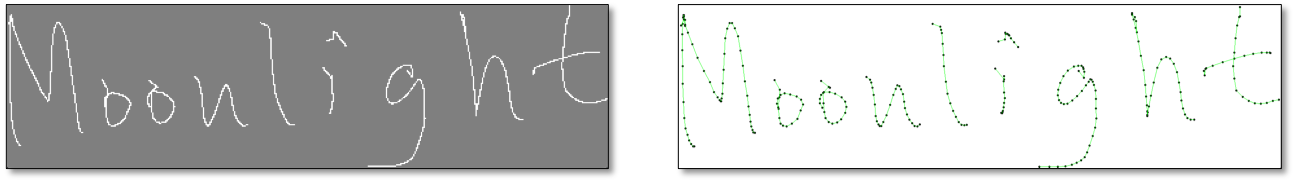}
		\caption{Signature as a sequence of stroke points}
		\label{figure:example_signature}
	\end{centering}
\end{figure}

Raw signature instances, essentially handwritten texts, are collected by recording the motions of digital pens on a tablet. They can be regarded a sequence of strokes which consist of a series of stroke points, as shown in Figure~\ref{figure:example_signature}. For a more rigorous description, denote $S = \{s_1, s_2, \cdots , s_n\}$ as a handwritten signature instance, making up by a sequence of strokes, where $s_i$ refers as the $i$-th stroke in current handwritten signature instance. One stroke $s$ comprises a set of sequential stroke points, i.e., $s = \{(x_1, y_1, p_1, t_1), (x_2, y_2, p_2, t_2), \cdots\}$, where $(x_j,y_j,p_j, t_j)$ is $j$-th stroke point with $(x_j,y_j)$ as its coordinate axis, $p_j\in\{0,1\}$ as pen up/down status to indicate the starting or ending stroke point, and $t_j$ as timestamp. 

Since the target signature data is sequential, tackling the verification task by recurrent neural network seems a natural choice. However, digital ink representations of the same handwritten text, especially the number of stroke points, may be varied dramatically, under different underlying ink representation algorithms, and may further cause significant generalization discrepancy. To solve this issue, rendering stroke points into binary images seems a reasonable solution, while lots of useful information like order and evolution of stroke points are lost during rendering. In order to balance rendering technique and sequential property of handwritten data, we make use of Path Signature Features (PSF) which has been demonstrated the effectiveness in online handwritten text recognition.


\subsection{PSF Feartures}
\label{sect:psf_old}

Path signature feature is defined over a path of stroke points, and usually calculated up to 2nd order~\cite{chen2017compact,ji2019generative}. More specifically, for each smallest stroke point path $\{(x_i,y_i), (x_{i+1}, y_{i+1})\}$ consisting of 2 consecutive points, the signature feature $p_{i,i+1}=(p_{i,i+1}^{(0)}, p_{i,i+1}^{(1)}, p_{i,i+1}^{(2)})\in\mathbb{R}^7$ is calculated as follows: 
\begin{equation}\label{eq.def:psf}
\begin{split}
p_{i,i+1}^{(0)}&=1\in \mathbb{R} \\
p_{i,i+1}^{(1)}&=(x_{i+1},y_{i+1})-(x_i, y_i)\in \mathbb{R}^2\\
p_{i,i+1}^{(2)}&=p_{i,i+1}^{(1)}\otimes p_{i,i+1}^{(1)}\in \mathbb{R}^4
\end{split}
\end{equation}
where $\otimes$ represents Kronecker matrix product. It follows the definition of PSF~\eqref{eq.def:psf} that PSF encodes geometrical and order information of stroke points into seven path signature maps to form a 3D tensor $T\in\mathbb{R}^{7\times H\times W}$, where $H$ and $W$ represent the height and with respectively. As~\cite{chen2017compact,ji2019generative}, we fix the height $H$ as 128, and let width be flexible.

\subsection{Temporal Enhanced PSF Features}
\label{sect:psf_new}

\begin{figure}[h]
	\begin{centering}
		\includegraphics[width=0.5\textwidth]{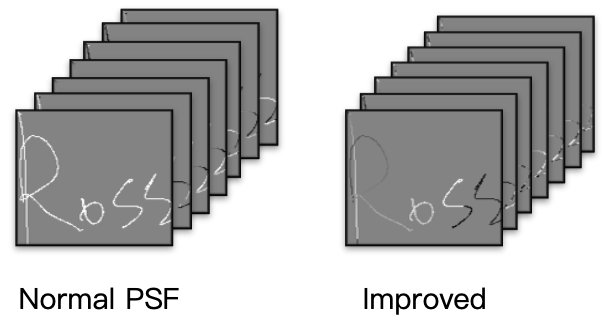}
		\caption{Comparison of two kinds of PSF}
		\label{figure:psf_comparison}
	\end{centering}
\end{figure}

The original PSF features in form of~\eqref{eq.def:psf} only encodes spatial information and order of stroke point into 3D tensor, which has been demonstrated its utility in Handwritten Recognition and Synthesis domain~\cite{chen2017compact}. However, remark here that the original PSF does not consider any temporal information of digital ink, e.g., the speed of pen movement, which may be crucial for many other applications. Particularly, for our target signature verification, the temporal information may be beneficial to achieve a better generalization performance, because a forger may spend longer time to mimic one genuine signature. Inspired by it, to incorporate temporal information, we design an enhanced PSF features as follows:

\begin{equation}\label{eq.def:temporalpsf}
\begin{split}
\tau_i & =\ln{(t_i+1)}\\
\tilde{p}_{i,i+1}^{(0)}&=\tau_ip_{i,i+1}^{(0)}=\tau_i\in \mathbb{R} \\
\tilde{p}_{i,i+1}^{(1)}&=\tau_ip_{i,i+1}^{(0)}=\tau_i(x_{i+1},y_{i+1})-\tau_i(x_i, y_i)\in \mathbb{R}^2\\
\tilde{p}_{i,i+1}^{(2)}&=\tau_i^2p_{i,i+1}^{(0)}=\tau_i p_{i,i+1}^{(1)}\otimes \tau_i p_{i,i+1}^{(1)}\in \mathbb{R}^4
\end{split}
\end{equation}
where a coefficient $\tau$ is computed as the logarithm of i-th stroke point's timestamp, then each PSF feature point is multiplied by this coefficient correspondingly. By utilizing such modified PSF features, the temporal information is involved as illustrated in Figure~\ref{figure:psf_comparison}. The effectiveness of such temporal enhanced PSF will be verified later. 

\section{Deep Learning Methods}\label{sect:nn_methods}

In this section, we present varying deep learning model architectures that utilizes the PSF features described in Section~\ref{sect:feature_extraction} for signature verification.

\begin{itemize}

\item \textbf{Convolutional Neural Networks}

The PSF features are essentially a 3D tensor, i.e., $T\in\mathbb{R}^{7\times H\times W}$. To handle such 3D tensors, Convolutional neural network (CNN) is a natural choice which have demonstrated its remarkable success in plentiful computer vision and handwritten applications~\cite{yang2015chinese}. Among numerous popular CNNs, we make use of LeNet-5~\cite{el2016cnn}, a classic CNN architecture for handwritten character recognition as the basic model, and adjust its structure properly to feed PSF tensors and binary classification as shown in Figure~\ref{figure:lenet5}.

\begin{figure}[h]
	\begin{centering}
		\includegraphics[width=1\textwidth]{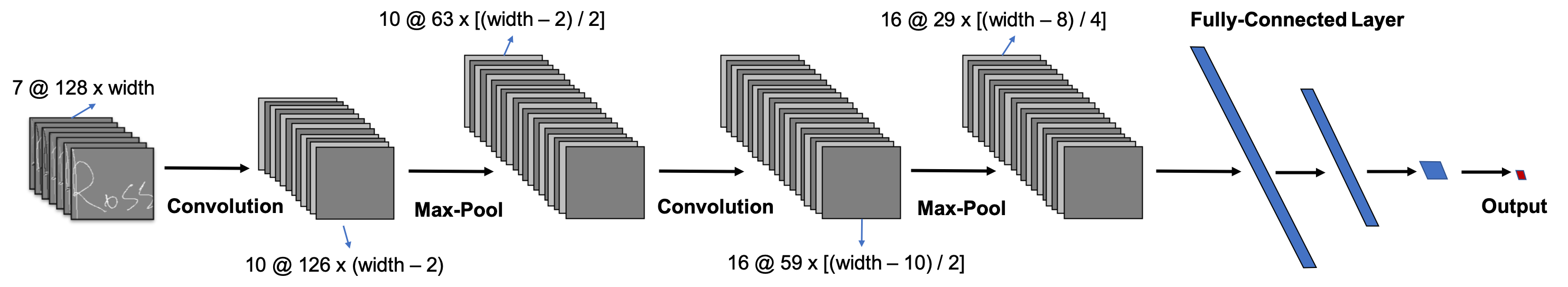}
		\caption{LeNet-5 Structure}
		\label{figure:lenet5}
	\end{centering}
\end{figure}

Since the different signatures possess varying lengths, which causes the extracted PSF tensors also own different widths, see Figure~\ref{figure:differentsignatures}. The different widths further result in the discrepancy on input sizes of fully connected layers. To address this problem, in the basic CNN model, we scale PSF tensors to fix the width as its height, namely 128. 

\begin{figure}[h]
	\begin{centering}
		\includegraphics[width=0.7\textwidth]{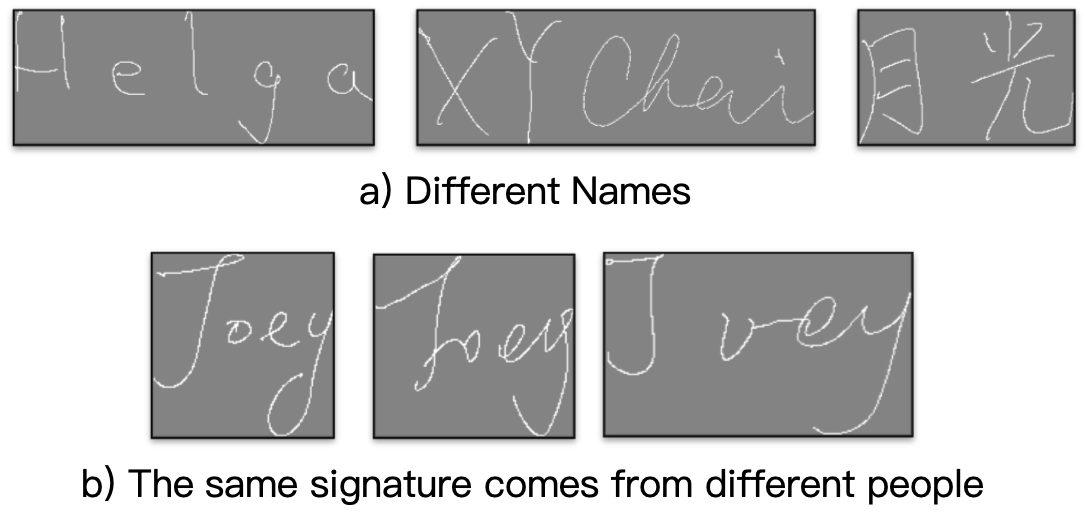}
		\caption{Different Signatures and Different Widths}
		\label{figure:differentsignatures}
	\end{centering}
\end{figure}

\item \textbf{Reccurent Neural Network}

Signatures consist of a sequence of strokes, of which is composed of a sequence of stroke points. For such sequential data, Recurrent Neural Network (RNN) seems valuable model architecture as well. However, RNN has a drawback of overflow as the length of input sequential data is enormous. To avoid this issue, we resample each signature to represent it by using the same small number of stroke points. Then we construct a RNN as shown in Figure~\ref{figure:RNN} where the last output of the RNN cell serves as the probability of genuine and forgery.  

\begin{figure}[h]
	\begin{centering}
		\includegraphics[width=1\textwidth]{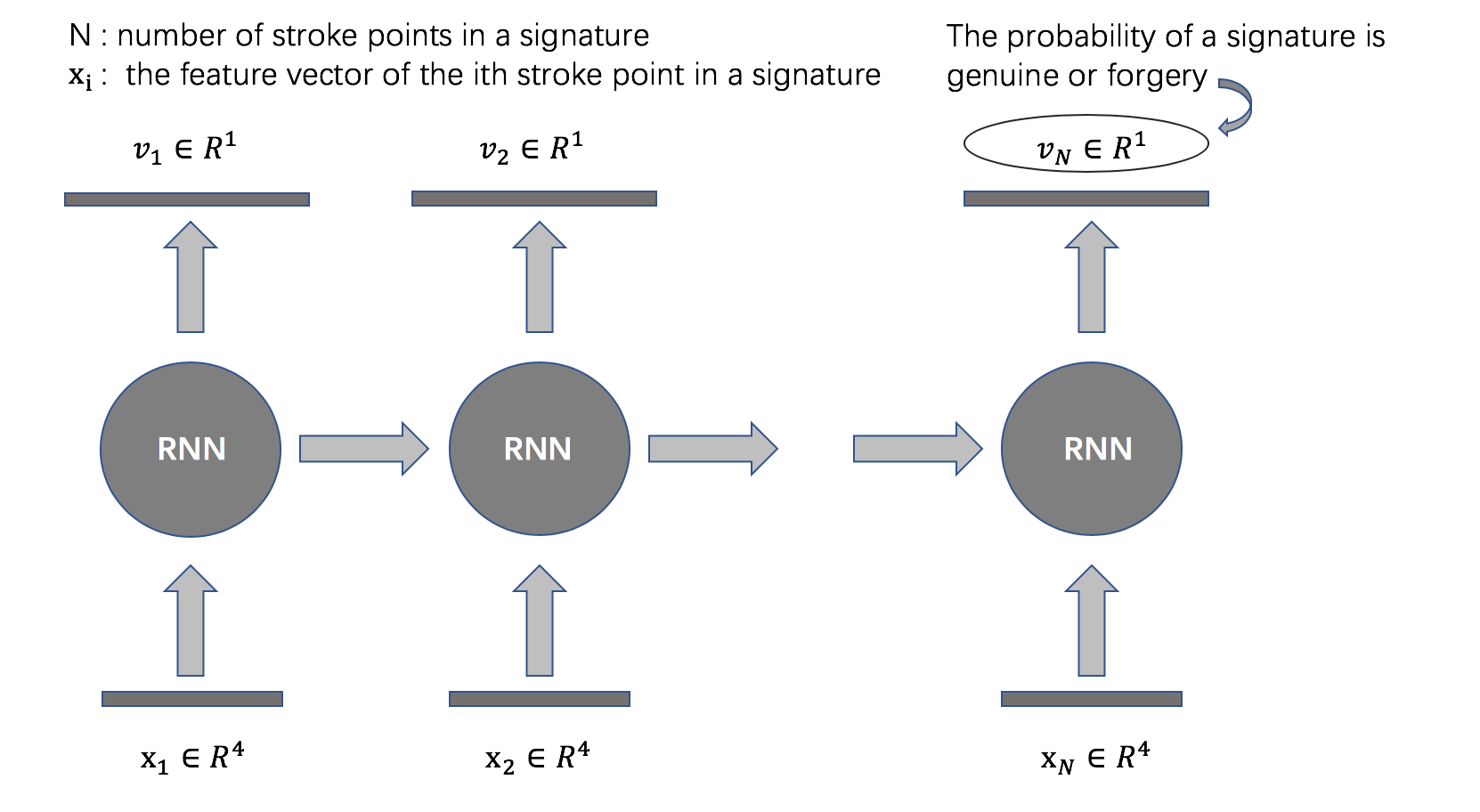}
		\caption{RNN Structure}
		\label{figure:RNN}
	\end{centering}
\end{figure}

\item \textbf{CNN-LSTM Compact Model}

Scaling and fixing the width and height of PSF tensors has the drawback that breaks down the original geometrical property of signature data.  To address this issue, we construct a CNN-LSTM compact model to allow flexible width input and achieve effective binary classification simultaneously. 

At first, the seven path signature maps extracted as described above are stacked to form a seven-channel PSF tensor denoted as $S$, which serves as the input of the following CNN-LSTM binary classification model, as illustrated in Figure~\ref{figure:cnn-lstm}. The CNN-LSTM model consists of a CNN and an LSTM, whose configurations are displayed in Table~\ref{tab:cnn-lstm}. CNN serves to encode extracted path signature feature into a matrix, since the input 3D tensor may possess varying widths, consequently results in that the output matrix of CNN possesses varying columns. Then each column of this encoded matrix is further fed into an LSTM cell sequentially, an FNN is assembled on the top of last output of the LSTM to calculate the probability of current instance as genuine written text. The whole procedure is shown in Figure~\ref{figure:cnn-lstm}.

\begin{figure}[h]
	\begin{centering}
		\includegraphics[width=1.0\textwidth]{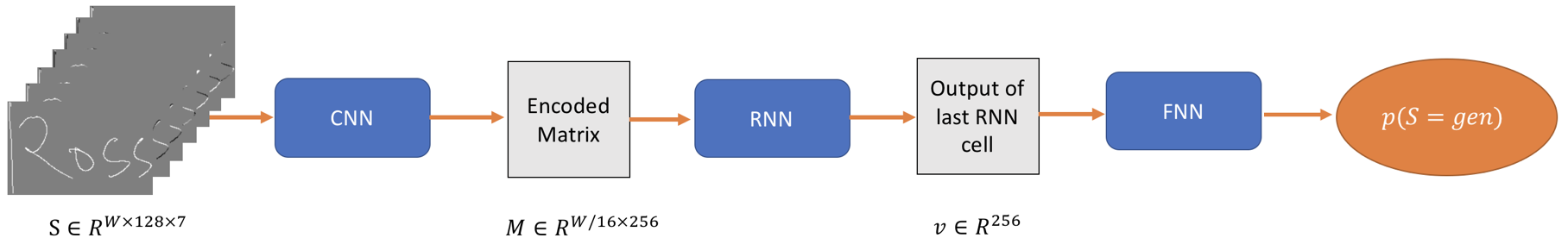}
		\caption{CNN-LSTM Architecture}
		\label{figure:cnn-lstm}
	\end{centering}
\end{figure}

\begin{table}[h]
	\center
	\caption{Architecture Configuration of CNN-LSTM Model, where input shape, kernel shape and output shape are in the manners of (width, height, channel), (\# of kernels, mask, stride) and (width, height, channel) respectively. The stride can either be a single number for both width and height stride, or a tuple (sW, sH) representing width and height stride separately.}
	\begin{tabular}{|c|c|c|c|c|}
		\hline
		& Input Shape   & Kernel Shape          &     Output Shape & Hidden Size  \\ \hline
		Conv1   &  $(W_I, 128, 7)$  & $(32, 3, 1)$        &  $(W_I, 128, 32)$   &   -- \\ \hline
		AvgPool   &  $(W_I, 128, 32)$  & $(-, 2, 2)$        &  $(W_I/2, 64, 32)$   &   -- \\ \hline
		Conv2   &  $(W_I/2, 64, 32)$  & $(64, 3, 1)$        &  $(W_I/2, 64, 64)$   &   -- \\ \hline
		AvgPool   &  $(W_I/2, 64, 64)$  & $(-, 2, 2)$        &  $(W_I/4, 32, 64)$   &   -- \\ \hline
		Conv3   &  $(W_I/4, 32, 64)$  & $(128, 3, 1)$       &  $(W_I/4, 32, 128)$   &   -- \\ \hline
		Conv4   &  $(W_I/4, 32, 128)$  & $(256, 3, (1, 2))$        &  $(W_I/4, 16, 256)$   &   -- \\ \hline
		AvgPool   &  $(W_I/4, 16, 256)$  & $(-, 2, 2)$      &  $(W_I/8, 8, 256)$   &   -- \\ \hline
		Conv5   &  $(W_I/8, 8, 256)$  & $(128, 3, (1, 2))$       &  $(W_I/8, 4, 128)$   &   -- \\ \hline
		Conv6   &  $(W_I/8, 4, 128)$  & $(256, 3, (1, 2))$       &  $(W_I/8, 2, 256)$   &   -- \\ \hline
		AvgPool   &  $(W_I/8, 2, 256)$  & $(-, 2, 2)$       &  $(W_I/16, 1, 256)$   &   -- \\ \hline
		
		LSTM     & 256 & -- & -- & 256 \\\hline
		FC1      & 256 & -- & --  & 128 \\\hline
		FC2      & 128 & -- & --  & 1 \\\hline  
	\end{tabular}
	\label{tab:cnn-lstm}
\end{table}

\end{itemize}

\section{Numerical Experiments}\label{sec.numericexp}

In this section, we present and discuss experimental results of the constructed models in Section~\ref{sect:nn_methods} to numerically demonstrate the effectiveness of some built model candidates. We at first describe the dataset in Section~\ref{sec:dataset}, followed by presenting the experimental settings in Section~\ref{sec.expsetting}, and show numerical results in Section~\ref{sec.results}. 

\subsection{Dataset}\label{sec:dataset}
The training and testing datasets is selected as the public benchmark signature data, namely SVC2004, which is used in the first world signature recognition competition \cite{meshoul2010novel}. These signature samples are collected through digital input boards and related equipments, not only providing signature information but also collecting information related to intrapersonal features, such as writing speed and pressure.  SVC2004 contains two tasks, where each task collects 40 sets and each set contains 20 real signatures signed by one person and 20 signatures by at least four others who imitated the handwritten signature. For training and testing purposes, we split the whole dataset randomly into 80 percentage as training and remaining 20 percentage as testing. Each signature instance is represented as a series of stroke points. Each stroke point consists of several features, i.e., index, x and y coordinate, timestamp, pen up/down status,  azimuth, altitude, and pressure. Since only the former four features are included in both tasks, we discarded the last three features in our experiments.



\subsection{Experimental Settings}\label{sec.expsetting}
The mini-batch size of our experiment is set as 10. We make use of standard cross entropy loss,  and select Adam optimizer to minimize the selected objective function.
\begin{figure}[h]
	\begin{centering}
		\includegraphics[width=1\textwidth]{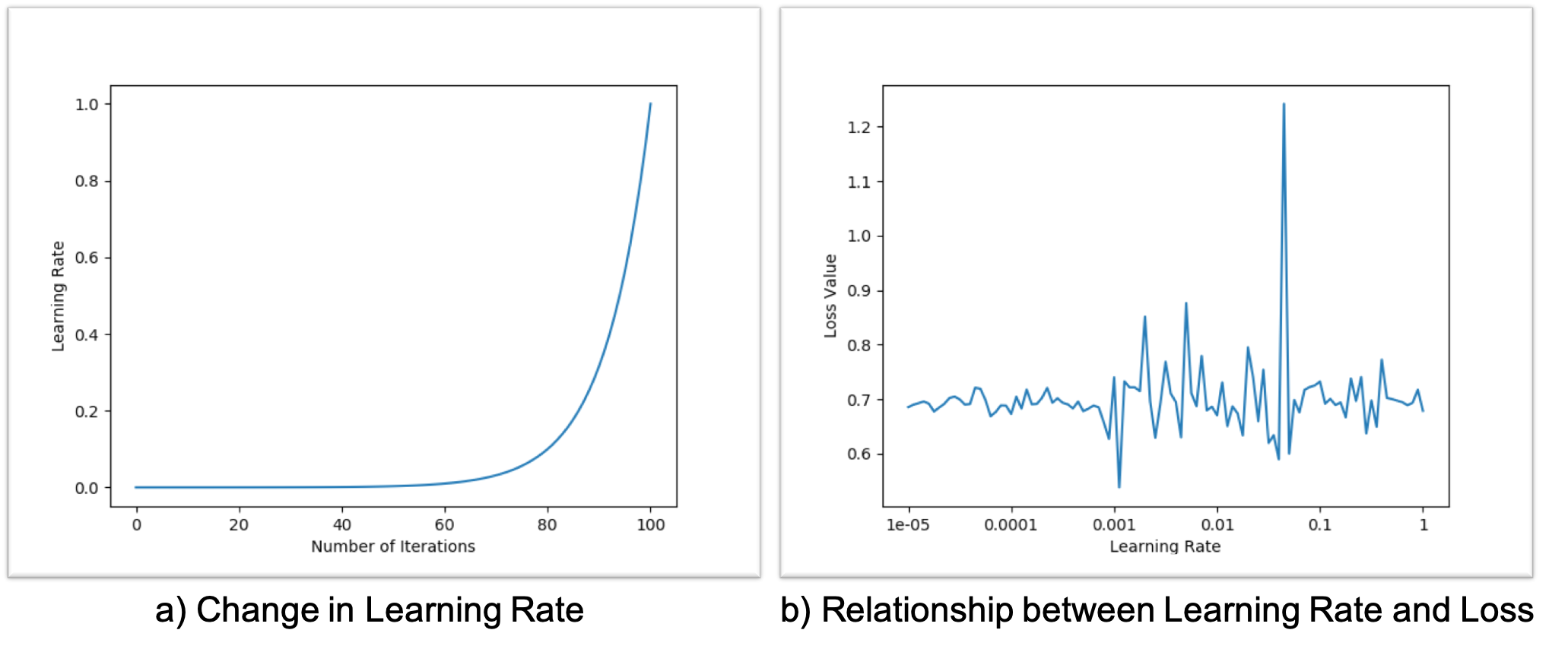}
		\caption{Learning Rate Slection Process}
		\label{figure.LR}
	\end{centering}
\end{figure}

Learning rate plays an vital role in the deep learning applications. There are significant numbers of methods to find optimal initial learning rate efficiently. In order to find an optimal initial learning rate, we use the method in~\cite{smith2017cyclical}.  Particularly, at first, we set a minimal learning rate, and after each batch, increase the learning rate exponentially. Then, looking for the lowest point of loss, and select the initial learning rate as the one corresponding to the lowest loss function value, as shown in  Figure.~\ref{figure.LR}. Such initial learning rate selection mechanism helps our training process a lot. After determining it, we decay the learning rate exponentially per epoch by a ratio as 0.95.

\subsection{Numerical Results}\label{sec.results}

We now summarize the generalization performances of varied deep learning model architectures on testing dataset in Table~\ref{table:results}. To facilitate comparison, the best values are marked as red, and the second best values are marked as blue.  The ``old PSF" refers the original PSF in form of~\eqref{eq.def:psf},  the ``new PSF" refers to temporal enhanced PSF as~\eqref{eq.def:temporalpsf}, and the ``stacked PSF" means a tensor formed by stacking original PSF and temporal enhanced PSF along channel axis. 

\begin{table}[]
	\centering
	\caption{Classification results on testing dataset of varying model architectures. }
	\begin{tabular}{lcccccc}
		\hline
		& \multicolumn{2}{c}{CNN} &  & \multicolumn{3}{c}{CNN-RNN}   \\
		& old PSF & new PSF & RNN &  old PSF & new PSF & stacked PSF\\  \hline
		Accuracy	&  0.514 & 0.739 & 0.728 & 0.738 & \textcolor{blue}{0.842} & \textcolor{red}{0.903}   \\
		Precision	 &  0.508 & 0.728 & 0.678 & 0.721 & \textcolor{blue}{0.892} & \textcolor{red}{0.930}  \\
		Recall	       & \textcolor{red}{0.925}  & 0.762 & 0.712  & 0.775 & 0.778 & \textcolor{blue}{0.872}  \\
		F1-score     & 0.656 & 0.745 & 0.694  & 0.747 & \textcolor{blue}{0.831} & \textcolor{red}{0.900}  \\
		\hline
	\end{tabular}
	\label{table:results}
\end{table}

\begin{itemize}
	
	\item \textbf{CNN with original PSF}: As shown in Table~\ref{table:results}, CNN with original scaled PSF features as input can not achieve satisfactory classification performance. More specifically, as illustrated in Figure~\ref{figure:cnn_loss}, after numerous training epochs, the value of loss function converges to a high value, namely 0.693. Such bad convergence results in a 51.4\% accuracy, high recall but low precision on testing datasets, which means  that the model tends to judge most of the data points as positive instances. This result also indicates that even though PSF in form of~\eqref{eq.def:psf} serves as very effective features for handwritten recognition, it helps limited in signature verification. The reason is because of the low interpersonal variance, i.e., if someone is aware of your signature ahead, he/she can probably mimic a very similar forgery signature in terms of spatial property. Due to the high spatial similarity, the original PSF as~\eqref{eq.def:psf} does not possess the capability to distinguish the discrepancy. Therefore, the original PSF is not an effective feature for signature verification, and needs to be improved. This leads to our subsequent algorithm improvement and the next series of experiments.
	
	\begin{figure}[h]
	\begin{centering}
		\includegraphics[width=1\textwidth]{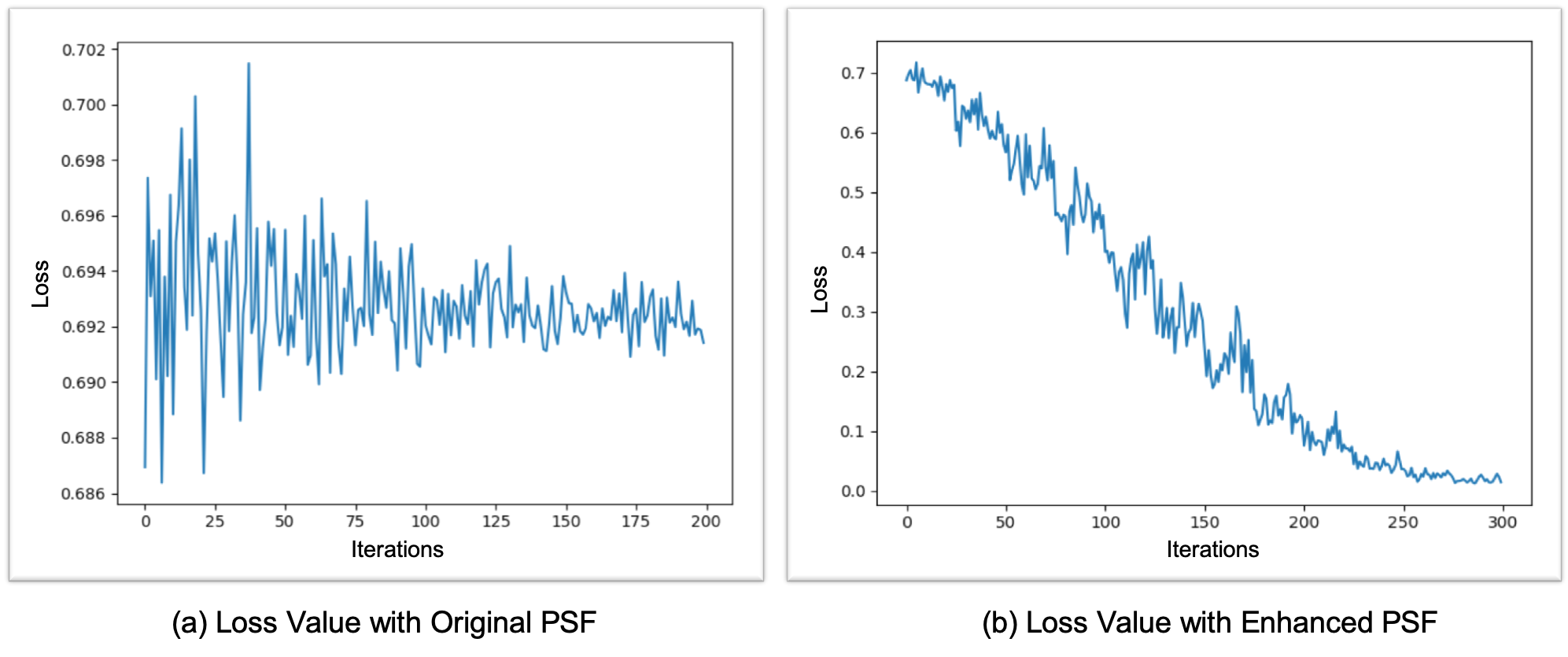}
		\caption{Loss Value in CNN Experiments}
		\label{figure:cnn_loss}
	\end{centering}
\end{figure}	

	\item \textbf{CNN with enhanced PSF}: After improving the PSF by incorporating temporal information into the tensor entries, we feed such enhanced PSF into the same CNN structure, and observe an significant improvement shown in Table~\ref{table:results}. Particularly, regarding the evolution of loss function values, we can see as the number of epochs increases, the loss function values keeps decreasing generally and stalks around 0.01. On the other hand, the accuracy also rises to 73.9\%, with relatively high recall as 76.2\% and precision as 72.8\%. It is a strong evidence that our temporal enhanced PSF works well to encode and enlarge the temporal discrepancy between genuine and forgery signatures.

	\item \textbf{RNN}: Before employing RNN on signature verification, we at first resample the stroke points of each signature instance to let them share the same number of stroke points by  uniform sampling algorithm in~\cite{wolin2008shortstraw} with some modifications. LSTM is selected the RNN cell, and then we can achieve classification metrics around 60-70\%.  
		
	\item \textbf{CNN-RNN Compact Model with original PSF}: We then use the CNN-LSTM model to allow original PSF tensors with varying widths. As shown in Table~\ref{table:results}, the performance of the CNN-LSTM model exceeds the ordinary CNN a lot, and is slightly better than the RNN model. The accuracy, precision, recall and F1 score increase to 70.0-80.0\%, from which we can see CNN-LSTM architecture is a more effective topological structure to achieve better generalization performance.

	\item \textbf{CNN-RNN Compact Model with enhanced PSF}: Previous experiments have revealed that both the temporal enhanced PSF and CNN-LSTM structure are beneficial for signature verification. Therefore, we expect that the combination of the improved PSF and CNN-LSTM models may achieve more outstanding results. The experimental results confirmed our hypothesis that this accuracy rate rose to 84.3\%, which is the best result so far.

	\item \textbf{CNN-LSTM Compact Model with stacked PSF}: To seek further accuracy gain, we design relative aggressive features by stacking the original PSF and temporal enhanced PSF together to get a new tensor with number of channels as 14. Besides, we also made changes in the neural network part, such as dropout technique. Finally, after some minor parameter adjustments, we got an accuracy rate of 90.3\%, precision of 93.0\%, recall as 87.2\%, as presented in Table~\ref{table:results}.
	
\end{itemize}

\section{Conclusion And Discussion}\label{sec.conclusion}

In this paper, we constructed a few modern deep learning methods to achieve signature verification, including CNN, RNN and CNN-RNN compact architectures.  We make use of Path Signature Feature as input to encode spatial information of signature instances, and improve original PSF with incorporated temporal information to resolve the issue of low interpersonal variance.  Our numerical experiments demonstrate the effectiveness of our model architecture and improved PSF features.



\end{document}